\documentclass{INTERSPEECH2023}

\usepackage{CJKutf8,booktabs,multirow}
\title{A Lexical-aware Non-autoregressive Transformer-based ASR Model}
\name{Chong-En Lin, Kuan-Yu Chen}
\address{
  National Taiwan University of Science and Technology, Taiwan}
\email{celin@nlp.csie.ntust.edu.tw, kychen@mail.ntust.edu.tw}

\begin{document}

\maketitle
 
\begin{abstract}
Non-autoregressive automatic speech recognition (ASR) has become a mainstream of ASR modeling because of its fast decoding speed and satisfactory result. To further boost the performance, relaxing the conditional independence assumption and cascading large-scaled pre-trained models are two active research directions. In addition to these strategies, we propose a lexical-aware non-autoregressive Transformer-based (LA-NAT) ASR framework, which consists of an acoustic encoder, a speech-text shared encoder, and a speech-text shared decoder. The acoustic encoder is used to process the input speech features as usual, and the speech-text shared encoder and decoder are designed to train speech and text data simultaneously. By doing so, LA-NAT aims to make the ASR model aware of lexical information, so the resulting model is expected to achieve better results by leveraging the learned linguistic knowledge. A series of experiments are conducted on the AISHELL-1, CSJ, and TEDLIUM 2 datasets. According to the experiments, the proposed LA-NAT can provide superior results than other recently proposed non-autoregressive ASR models. In addition, LA-NAT is a relatively compact model than most non-autoregressive ASR models, and it is about 58 times faster than the classic autoregressive model.
\end{abstract}

\section{Introduction}

Automatic speech recognition (ASR) is one of the important research in the context of natural language processing, where the goal is to convert a speech into a corresponding text sequence. In recent years, end-to-end based ASR \cite{LAS,hybrid_ctc/attention,CTC} has created a dominant paradigm, and the school of research can be further divided into autoregressive (AR) and non-autoregressive (NAR) manners according to different decoding philosophies. AR models predict each token by considering all the previous historical information, can be combined with a language model to enhance the performance easily, and usually achieve acceptable performances. However, they often struggle with the decoding speed barrier caused by autoregressive one-by-one decoding. Orthogonal to the AR models, NAR methods assume that all the tokens are conditionally independent, so the decoding results can be generated in parallel at once. In short, the design allows NAR models to deliver faster-decoding speeds in practice, but their recognition performance is usually only comparable to AR models.

In order to boost performance, various research has been devoted to improving NAR ASR. A summary of these models shows two main directions for improvement. On the one hand, some studies concentrate on relaxing the conditional independence assumption of the learning objectives from a theoretical perspective \cite{CCTC,CAKT}. As a result, the improved models are expected to equip the ability to take contextual information into account during decoding. On the other hand, many researchers turn to leverage pre-trained speech and/or language models for building the NAR models. By distilling the knowledge or directly concatenating the pre-trained models to make the resulting ASR models inherit the merits of these pre-trained models \cite{CTC_KnowledgeTransfer,PLM-NAR-ASR,NAR_CTC/attention}. Although classical NAR techniques have progressed with these developments, they usually have a large set of model parameters and are impractical for real-world applications.

Aside from relaxing the conditional independence assumption and using pre-trained models, we propose a framework to enhance NAR ASR by explicitly making the model aware of the lexical information. Accordingly, a lexical-aware non-autoregressive Transformer-based ASR (LA-NAT) model, which consists of an acoustic encoder, a speech-text shared encoder, and a speech-text shared decoder, is introduced. The acoustic encoder is used to process the speech input and produce a set of acoustic features as usual. In order to make the ASR model learn the lexical information naturally, our idea is to construct a shared network for training text and speech data simultaneously. As a result, the model is expected to equip the linguistic knowledge because it must be able to mitigate text-oriented tasks. At the same time, the model is also used for speech recognition, so it would be a lexical-aware ASR model. The speech-text shared encoder and decoder are designed to make the idea work. As the proposed LA-NAT does not cascade large-scaled pre-trained models, its lightweight property is yet another advantage. Extensive experiments are conducted on the AISHELL-1 \cite{Aishell-1}, CSJ \cite{CSJ}, and TEDLIUM 2 \cite{TED-LIUM} datasets. The proposed LA-NAT not only demonstrates superior results on these corpora as expected but also reveals the lightweight property compared with other complicated models and achieves about $58$ times faster than the classic autoregressive model.

\section{Related Work}
Autoregressive (AR) ASR models aim to generate an output token sequence $Y=\{y_1,…,y_L\}$ for a given speech utterance $O$ by referring to $P(Y|O)$. According to the chain rule, a common simplification strategy is to decompose $P(Y|O)$ into a series of conditional probabilities, as
\begin{equation} 
    \label{eq: AR}
    \begin{aligned}
        Y^* &= \arg\max_{Y} P(Y|O)=P(y_1|O) \prod_{l = 2}^{L}P(y_l|y_{<l},O),
    \end{aligned}
\end{equation}
where $y_{<l}$ denotes the partial token sequence before $y_l$. Obviously, AR models are designed to generate tokens by considering all previously generated ones and acoustic statistics. Even though the philosophy promotes these models to take plenty of clues into account for making accurate predictions, there are usually several associated challenges. First, the left-to-right nature of the AR models limits the efficiency of the parallel computation and increases the difficulty of enhancing the inference speed. Second, during training, ground truth history tokens are used to predict the next one, but at the inference stage, only the generated sequence can be used as a condition to predict the upcoming token, which may contain recognition errors. This mismatch problem between training and inference is a common challenge for AR models \cite{Scheduled_Sampling}. A few such representative models are Listen, Attend and Spell (LAS) \cite{LAS}, Self-attention Transducer (SA-T) \cite{SA-T}, hybrid CTC/attention architecture \cite{hybrid_ctc/attention}, Speech-Transformer \cite{Speech-transformer}, and Memory equipped Self-attention (SAN-M) \cite{SAN-M}.

Non-autoregressive (NAR) ASR models assume that each token is conditionally independent of others that are given a speech signal. Formally, for a given speech utterance $O$, the NAR modeling decodes the corresponding token sequence $Y$ by referring to:
\begin{equation} 
    \label{eq: NAR}
    \begin{aligned}
        Y^* &= \arg\max_{Y} P(Y|O)=\prod_{l = 1}^{L}P(y_l|O).
    \end{aligned}
\end{equation}
Compared with \eqref{eq: AR}, without the dependency between the current and previous tokens, the NAR modeling can generate tokens in parallel to achieve low latency. Besides, the mismatch issue between training and inference no longer occurs. Despite their apparently much simpler designs, more and more NAR-based ASR models demonstrate rapid inference speed and competitive performances with conventional AR methods. In spite of the advantages NAR models seem to have over AR models, a major challenge of NAR modeling is determining the output length. Specifically, during inference, AR modeling stops generating tokens when an end-of-sentence token (e.g., $<$EOF$>$) is produced, while the NAR modeling requires additional efforts to post-process the predicted result \cite{CTC, LFML, Mask_CTC} or to estimate/adjust the length of the recognition sequence \cite{LASO_journal, Improved_Mask-CTC}. A few of the widely used models include Connectionist Temporal Classification (CTC) \cite{CTC}, Listen and Fill in the Missing Letters (LFML) \cite{LFML}, Listen Attentively, and Spell Once (LASO) \cite{LASO_journal}, Pre-trained Language Model-based NAR ASR (PLM-NAR-ASR) \cite{PLM-NAR-ASR} and Mask CTC \cite{Mask_CTC}. Comprehensive comparisons between AR and NAR ASR models are found in the literature \cite{NAT_ASR_CTC, Compare_NAR, Pushing_NAR-ASR}.

\section{Proposed Method}

The proposed LA-NAT is composed of an acoustic encoder, a speech-text shared encoder, and a speech-text shared decoder. The model architecture is illustrated in Figure 1.

\subsection{Acoustic Encoder}
\label{sec:enc}
Each training example consists of a speech signal $O$ and its corresponding token sequence $Y$, where the speech signal is quantized into a series of $T’$ acoustic feature frames $\{o_1,...,o_{T'}\}$ and $Y$ denotes a sequence of $L$ tokens $\{y_1,...,y_L\}$. The acoustic encoder uses two layers of the 2D convolutional neural networks to encode temporal and spectral vicinity statistics of acoustic feature frames and downsample the resolution in time. Next, the sinusoidal positional embedding is added for each feature to retain the order in the line \cite{Attention,BERT}. Afterward, we stack a series of $N_a$ Transformers to reformulate the acoustic characteristics because the self-attention mechanism can blend short-term and long-term information together. Consequently, a set of acoustic features $H^{O}=\{h_1^{O},…, h_{T}^{O}\} \in R^{d×T}$ is derived. A layer normalization layer, a linear layer, and a softmax activation function are sequentially stacked on $H^{O}$, and a CTC objective $\mathcal{L}_\text{CTC}$ is used to guide the model training toward minimizing the differences between the prediction and the ground-truth.

\subsection{Speech-Text Shared Encoder}
\label{sec:SR}
In order to make the ASR model learn lexical information naturally, a meticulous speech-text shared module, including an encoder and a decoder, is introduced. To be more specific, for each training example, the acoustic encoder converts $O$ into a set of acoustic features $H^{O}$, while a simple table lookup is applied to translate ground truth text $Y$ into a series of token embeddings. Without loss of generality, the positional embedding is added for each token embedding to reveal the ordered information in line, and the set of representations is denoted as $H^{Y} \in R^{d×L}$. A speech-text shared encoder is employed to adjust the two modality features first, and then organize them into a set of representative information. Following the idea, a series of $N_{b}$ Transformer layers are applied to adjust acoustic- and text-level features (i.e., $H^{O}$ and $H^{Y}$), and a layer normalization layer is adopted to stable the distribution of statistics. Accordingly, two set of features $H_{se}^{O} \in R^{d×T}$ and $H_{se}^{Y} \in R^{d×L}$ are obtained for speech and text inputs, respectively.

Next, inspired by the memory network \cite{Memory}, we initialize $M$ memory slots to act as a set of anchors to rearrange and reorganize speech and text information. Concretely, the memory slots and a set of features (i.e., $H_{se}^{O}$ or $H_{se}^{Y}$ ) are fed into a Transformer, where the former is used to query the latter. Subsequently, $N_c$ Transformers, where the queries are output from the previous Transformer, while the keys and values are identical to the original set of features (i.e., $H_{se}^{O}$ or $H_{se}^{Y}$), are used to iteratively rearrange and reorganize the statistics to obtain a set of acoustic-based composite representations $H_{com}^{O}=\{h_{com_1}^{O},...,h_{com_M}^{O}\} \in R^{d×M}$ and a set of lexical-based composite representations $H_{com}^{Y}=\{h_{com_1}^{Y},...,h_{com_M}^{Y}\} \in R^{d×M}$. Since the content of speech and text input is the same, the resulting acoustic-based and lexical-based composite representations should be able to align one-to-one. Therefore, a contrastive objective $\mathcal{L}_\text{Cont}$ is used to maximize the matching degree between each pair of composite representations:
\begin{align}
\begin{aligned}
    \mathcal{L}_\text{Cont}=-\sum_{i=1}^{M}log\frac{e^{\text{cos}(H_{com_{i}}^{O},H_{com_{i}}^{Y})/\tau}}
    {\sum_{j=1}^{M}e^{\text{cos}(H_{com_{i}}^{O},H_{com_{j}}^{Y})/\tau}}\\
    -\sum_{j=1}^{M}log\frac{e^{\text{cos}(H_{com_{j}}^{O},H_{com_{j}}^{Y})/\tau}}
    {\sum_{i=1}^{M}e^{\text{cos}(H_{com_{j}}^{O},H_{com_{i}}^{Y})/\tau}},
\end{aligned}
\end{align}
where \text{cos( , )} denotes the cosine similarity function, and $\tau$ is a scaling factor. It is worth noting that memory slots are model parameters that should be updated by training. Besides, the memory network-liked procedure can not only deal with the length mismatch between acoustic- and text-level features (i.e., $H_{se}^{O}$ and $H_{se}^{Y}$) but also work as a set of pivots to align and capsule information from the two modalities of features.

\subsection{Speech-Text Shared Decoder}
\label{sec:DTE}

After the speech-text shared encoder, a speech-text shared decoder is performed at the last step. The decoder is employed to do speech recognition with acoustic-based representations, while it also has to reconstruct the original token sequence with lexical-based representations. In detail, for the former, we initialize a set of positional embeddings $H_{PE} \in R^{d×L}$ as queries to retrieve information from acoustic representations $[H_{se}^{O};H_{com}^{O}] \in R^{d×(T+M)}$ by using $N_d$ Transformer layers. Except for the first one Transformer, the queries are output from the previous Transformer, and the keys and values are always identical to the concatenated acoustic features $[H_{se}^{O};H_{com}^{O}]$. It should be emphasized that we use a fusion mask to constrain the attention scope in each Transformer \cite{CASS-NAT}. The Viterbi algorithm is used to match the CTC alignment with the ground truth into a trigger mask as the left half of the fusion mask, and elements in the right-hand side of the fusion mask are all set to one. By doing so, each query can capture local information from features produced by the speech-text shared encoder $H_{se}^{O}$ and global information from acoustic-based composite features $H_{com}^{O}$ by the cross-attention mechanism. In other words, $H_{se}^{O}$ is a set of low-level fine-grained acoustic features, while $H_{com}^{O}$ can be regarded as a set of coarse-grained auxiliary information. After that, the resulting features are passed through a sequence of $N_e$ Transformers to refine the statistics. A linear layer and a softmax activation function are sequentially stacked to construct an ASR model. Consequently, the cross-entropy loss $\mathcal{L}^{O}_\text{CE}$ is used to guide the model training toward minimizing the differences between the prediction and the ground-truth. During training, the teacher forcing strategy is used in our implementation, so the number of positional embeddings (i.e., $L$) is determined by referring to the ground truth.

Despite the speech part, the speech-text shared decoder is also designed to reconstruct the original token sequence by feeding lexical representations. Again, a set of positional embeddings is created to query the lexical-based statistics $[H_{se}^{Y};H_{com}^{Y}]$ and reorganize them with the corresponding attention scores by cross-attention mechanism. Except for the first Transformer, the queries are generated by the previous Transformer, and the keys and values are always identical to the concatenated features $[H_{se}^{Y};H_{com}^{Y}]$. Following, the resulting features are passed through a set of Transformers to manipulate the statistics, and then a linear layer and a softmax activation function are sequentially stacked. Finally, the cross-entropy loss $\mathcal{L}^{Y}_\text{CE}$ is used to guide the model training toward minimizing the reconstruction errors.

During inference, we use the CTC branch to calculate the fusion mask and decide the length of the recognition result. It is worth noting that text input is only used for training, and the proposed ASR model is performed in a non-autoregressive manner. In a nutshell, the proposed LA-NAT introduces a speech-text shared encoder and a speech-text shared decoder to train speech and text input simultaneously while with different tasks. Thereby, LA-NAT can learn linguistic knowledge from the text data so as to enhance the performance of ASR.

\begin{figure}[htb]
  \centerline{\includegraphics[width=8.5cm]{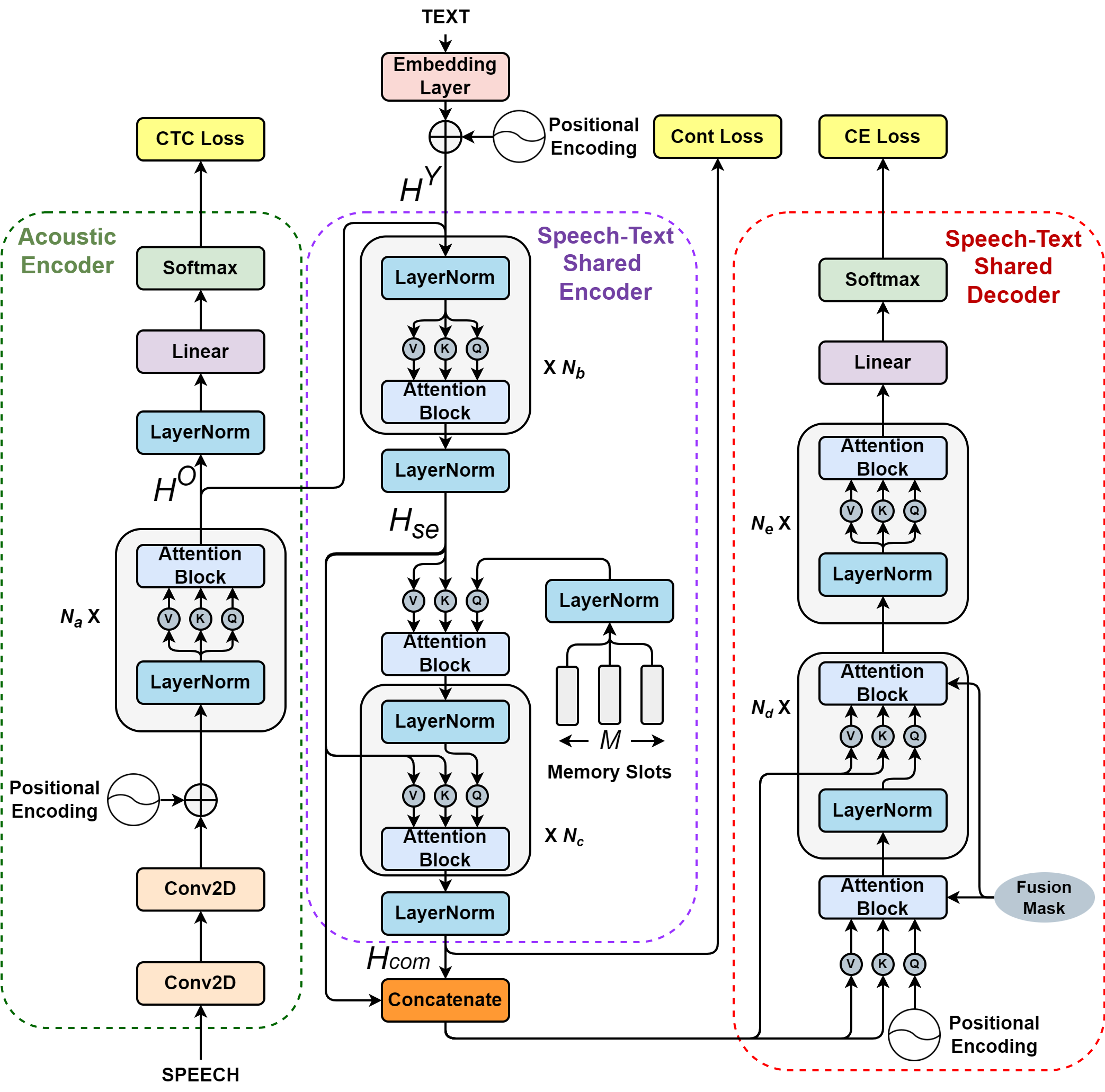}}
\caption{The model architecture of the proposed LA-NAT.}
\end{figure}

\subsection{Training Method}
\label{sec:SL}
Since the proposed LA-NAT has to face two modality features, the training procedure must be carefully designed to stable the model training and obtain a better model. 

\textbf{Step1:} First of all, we collect a large set of text data to train the speech-text shared encoder and decoder with the reconstruction task (i.e., the objective function is $\mathcal{L}^{Y}_\text{CE}$) only. By doing so, linguistic knowledge can be learned and stored in the speech-text shared encoder and decoder models.

\textbf{Step2:} Following, we optimize all the model parameters, except the set of word embeddings used for translating text input to a sequence of token embeddings, by using speech data only toward minimizing $\lambda_\text{CTC}\mathcal{L}_\text{CTC}+\lambda^{O}_\text{CE}\mathcal{L}^{O}_\text{CE}$ (cf. Sections 2.1 and 2.3). The hyper-parameters $\lambda_{CTC}$ and $\lambda^{O}_{CE}$ are set to $0.3$ and $1.0$, respectively. On top of the first stage, the speech-text shared encoder and decoder have been pre-trained with text data, so we expect the ASR model can be more robust base on the learned linguistic knowledge.

\textbf{Step3:} In the last step, both speech and text data are used to update the entire model. The training objective is to minimize all the loss functions $\lambda_\text{CTC}\mathcal{L}_\text{CTC}+\lambda_\text{Cont}\mathcal{L}_\text{Cont}+\lambda^{Y}_\text{CE}\mathcal{L}^{Y}_\text{CE}+\lambda^{O}_\text{CE}\mathcal{L}^{O}_\text{CE}$. The weighting factors $\lambda_\text{CTC}$, $\lambda_\text{Cont}$, $\lambda^{Y}_\text{CE}$, and $\lambda^{O}_\text{CE}$ are set to $0.3$, $1.0$, $0.3$, and $1.0$, respectively. 

It's interesting to note that the text data used in the first step can come from either the speech corpus, another selected collection, or a combination of them. Since the text data is easy to collect, the proposed LA-NAT can thus enjoy the advantage of inheriting linguistic knowledge from a large set of text data and becoming a lexical-aware Transformer-based NAR ASR model.

\begin{table*}[]
\caption{\label{tab:result}Experimental results on AISHELL-1 (CER\%), CSJ (CER\%) and TEDLIUM 2 (WER\%), respectively. The model parameters and real-time factors (RTF) are calculated by using the AISHELL-1 dataset.}
\centering
\scalebox{1}{
\begin{tabular}{clccccccccc}
\toprule
\multicolumn{2}{c}{\multirow{2}{*}{MODEL}} & \multicolumn{2}{c}{AISHELL-1} & \multicolumn{3}{c}{CSJ} & \multicolumn{2}{c}{TEDLIUM 2} & \multirow{2}{*}{\#Param.} & \multirow{2}{*}{RTF} \\ \cline{3-9}
\multicolumn{2}{c}{} & Dev & Test & Eval1 & Eval2 & Eval3 & Dev & Test &  &  \\ 
\hline
\hline
\multirow{5}{*}{AR} 
 & LAS \cite{LAS} & - & 8.7 & - & - & - & - & - & 156M &  -\\
 & SA-T \cite{SA-T} & 8.3 & 9.3 & - & - & - & - & - & - &  -\\
 & ESPnet (Transformer) \cite{ESPnet} & 6.0 & 6.7 & 6.37 & 4.76 & 5.4 & 12.6 & 10.2 & 58M & 0.362 \\
 & SAN-M \cite{SAN-M} & 5.7 & 6.5 & - & - & - & - & - & - &  -\\
 & CAT \cite{CAT} & - & 6.3 & - & - & - & - & - & - &  -\\
\hline
\multirow{10}{*}{NAR} 
 & ST-NAT \cite{Spike-triggered} & 6.4 & 7.0 & - & - & - & - & - & 31M &  -\\
 & Mask CTC \cite{Mask_CTC} & 6.0 & 6.7 & 6.56 & 4.57 & 4.96 & 11.9 & 10.7 & 29.7M &  -\\
 & LASO- Big \cite{LASO_journal} & 5.8 & 6.4 & - & - & - & - & - & 105.8M & 0.0042 \\
 & CTC \cite{Mask_CTC,Relax_CTC} & 5.7 & 6.2 & 6.51 & 4.71 & 5.49 & 12.8 & 12.2 & - & - \\
 & NAR-Transformer \cite{NAR_Transformer} & 5.3 & 5.9 & - & - & - & - & - & 29.7M & - \\
 & CASS-NAT \cite{CASS-NAT} & 5.3 & 5.8 & - & - & - & - & - & 33.2M & - \\
 & HANAT-Big \cite{HANAT} & 5.1 & 5.6 & - & - & - & - & - & 87M & - \\
 & Improved CASS-NAT \cite{Improved_CASS-NAT} & 4.9 & 5.4 & - & - & - & - & - & 38.3M & - \\
 & \textbf{LA-NAT Base} & \textbf{4.58} & \textbf{4.96} & \textbf{5.91} & \textbf{4.11} & \textbf{5.05} & \textbf{11.6} & \textbf{11} & 31.5M & 0.0062 \\
 & \textbf{LA-NAT Large} & \textbf{4.51} & \textbf{4.88} & \textbf{5.55} & \textbf{3.73} & \textbf{4.5} & \textbf{10.9} & \textbf{10.6} & 79.8M & 0.0067 \\
\bottomrule
\end{tabular}
}
\end{table*}

\section{Experiments}
\label{sec:exp}
\subsection{Experimental setup}
\label{sec:exp_setup}
The experiments are conducted on three datasets, including AISHELL-1, CSJ, and TEDLIUM 2. For pre-training, we use the AISHELL-2 \cite{Aishell-2}, the training set in CSJ, and LibriSpeech \cite{Librispeech} to initialize the speech-text shared modules. The lexicon sizes are $4,231$, $3,260$, and $5,000$ for Chinese, Japanese, and English, respectively. We use $80$-dimensional log Mel-filter bank features with $3$-dimensional pitch features, computed every 10ms with a window size of $25$ms, and the SpecAugment is applied for data augmentation. Speed perturbation is also used for AISHELL-1 and TEDLIUM 2. The hyper-parameters $\{N_{a}, N_{b}, N_{c}, N_{d}, N_{e} \}$ are set to $\{6,6,5,2,4\}$ for AISHELL-1, while $N_{a}$ is set to $12$ for CSJ and TEDLIUM 2. We build a base and a large models with hidden layer sizes $d$ of $256$ and $512$, respectively. Each CNN layer in the acoustic encoder has $256$ or $512$ filters, depending on whether it is a base or a large model, with a kernel size of $3$ and a stride of $2$. The intermediate dimension of the feedforward neural network is $2,048$ with $8$ attention heads, and the activation function is GLU. During the schedule learning, the model is trained with $40$ epochs in the first stage and $130$ epochs in the next two stages, and we average the model parameters of the last $10$ epochs as the final model. Each training batch consists of $100$ seconds of speech and accumulates gradients of $12$ steps.

\subsection{Experimental results}
\label{sec:exp_results}
Table 1 summarizes the experimental results of the proposed LA-NAT and several SOTA models. Information on the real-time factors (RTF) and the model sizes are also presented in Table 1. Several observations can be drawn from the results. At first glance, the results reveal that the NAR models have caught up to the efficiency and effectiveness of the AR ones, thereby demonstrating the potential of NAR ASR modeling. Second, the proposed LA-NAT can deliver better results and provide a faster decoding speed than the AR models. Taking LA-NAT Base and ESPnet (Transformer) as an example, regardless of the improvement in recognition performance, the former is not only nearly half the size of the latter, but the decoding speed is at least 58 times faster. Third, compared with NAR models, LA-NAT not only can give satisfactory results but also is a relatively exquisite model. Although LA-NAT only obtains competitive results on the TEDLIUM 2 dataset, we have proved the proposed LA-NAT can be done in both logographic (i.e., Chinese and Japanese) and alphabetic (i.e., English) languages. To further boost the performance, one of the emergent future works is to search for better configurations for alphabetic languages. A possible way is to leverage enhanced CTC-based objectives for training the acoustic encoder \cite{CCTC, CAKT, Relax_CTC}. Finally, it should be mentioned that the performance gap between the base and large models of the proposed LA-NAT is remarkable on the CSJ and TEDLIUM 2 corpora, but there is only a dearth of improvements on the AISHELL-1 dataset. A possible reason may be that both the CSJ and TEDLIUM 2 are more complicated in content than the AISHELL-1. In this case, scaling up the model size provides a straightforward way of increasing the capability and ability of the proposed LA-NAT.

\begin{table}[]
\caption{\label{tab:ablation1}Ablation study on AISHELL-1 dataset with only SpecAugment.}
\centering
\begin{tabular}{lcc}
\toprule
\multicolumn{1}{c}{Training Recipe} & Dev & Test \\ 
\hline
\hline
From scratch & 5.21 & 5.83 \\ 
Step 3 & 5.35 & 5.74 \\
Steps 1 and 3 & 5.22 & 5.62 \\
Steps 1, 2 and 3  & 4.84 & 5.33 \\ 
\bottomrule
\end{tabular}
\end{table}

\subsection{Ablation Studies and Analysis}
\label{sec:ablation}
To analyze the impact of lexical information in the proposed LA-NAT, we examine the results step by step. The experimental results are shown in Table 2. “From scratch” denotes LA-NAT is directly trained using a speech dataset without extra text corpus, pre-training, and text input. In other words, only the CTC loss $\mathcal{L}_\text{CTC}$ and cross-entropy loss for ASR $\mathcal{L}^{O}_\text{CE}$ are used to update the model parameters. It is thus a naïve baseline of LA-NAT. From the results, it is easy to conclude that linguistic knowledge can indeed help the ASR model. Furthermore, the well-designed training procedure can improve the effectiveness and efficiency of LA-NAT in learning and leveraging linguistic knowledge. The best setting (i.e., “Steps 1, 2 and 3”) can deliver up to $8.6\%$ relative CER reduction on the test set than the naïve setting (i.e., “From scratch”). In sum, the set of experiments stresses the utilities of the lexical information, the potential of the proposed LA-NAT, and the importance of the presented training procedure.

\section{Conclusions}
\label{sec:conclusion}
In this paper, we propose a novel lexical-aware non-autoregressive Transformer-based ASR framework, which indicates a potential way to enhance the performance of NAR ASR. The experimental results show that LA-NAT achieves competitive or SOTA results compared to other NAR models and is about 58 times faster than the classic AR model. In the future, we will continue improving the model architectures and exploring different training objectives and effective methods for modeling lexical-aware non-autoregressive ASR.

\vfill\pagebreak

\bibliographystyle{IEEEtran}
\bibliography{mybib}

\end{document}